\documentclass{article}

\usepackage{PRIMEarxiv}

\usepackage[utf8]{inputenc}
\usepackage[T1]{fontenc}
\usepackage{hyperref}
\usepackage{url}
\usepackage{booktabs}
\usepackage{amsfonts}
\usepackage{nicefrac}
\usepackage{microtype}
\usepackage{fancyhdr}
\usepackage{graphicx}
\usepackage{times}
\usepackage{natbib}         
\usepackage{caption}
\usepackage{subcaption}
\usepackage[section]{placeins}  

\usepackage{amsmath,amsfonts,bm}

\def\eqref#1{equation~\ref{#1}}

\def\1{\bm{1}}

\DeclareMathAlphabet{\mathsfit}{\encodingdefault}{\sfdefault}{m}{sl}
\SetMathAlphabet{\mathsfit}{bold}{\encodingdefault}{\sfdefault}{bx}{n}

\title{When Drift Detectors Cry Wolf: False Alarm Rates in Continuous ML Monitoring%
}

\author{
  Raj Shekhar Singh \\
  Indian Institute of Technology, Roorkee \\
  Roorkee, Uttarakhand, India \\
  \texttt{raj\_ss@ece.iitr.ac.in}
}

\begin{document}

\maketitle

\begin{abstract}
Drift detection is a core component of production machine learning monitoring systems, where detectors are used to compare incoming data with a reference distribution and trigger alerts when changes occur. However, these detectors are often evaluated in research settings that emphasize detection accuracy under synthetic shifts, while overlooking false alarms under continuous monitoring. In production environments, models are monitored repeatedly over time and across many features, and even small false positive rates can accumulate into frequent alerts, leading to alarm fatigue.

We empirically analyze false positive behavior across five commonly used drift detectors: PSI, KS, MMD, LSDD, and adversarial validation. Consistent with existing literature, PSI exhibits strong sensitivity to batch size, producing frequent false alarms at small sample sizes; however, we further observe that its behavior stabilizes and improves substantially once batch sizes exceed approximately 200 samples. In contrast, KS, MMD, and LSDD display persistent fluctuations across batch sizes, while remaining comparatively more reliable than PSI in low-data regimes. Applying a Bonferroni correction reduces false positive rates, but often at the cost of reduced true positive sensitivity, reinforcing the well-known stability--sensitivity trade-off in drift detection.

This work provides a systematic comparison of false positive behavior across multiple drift detectors under continuous monitoring conditions. We identify trade-offs across detector families and provide practical guidelines for selecting and calibrating drift detectors in production ML systems.
\end{abstract}


\section{Introduction}

Machine learning systems in production operate in environments that change over time. Shifts in user behavior, upstream pipelines, or data sources can alter the input distribution, leading to data drift and degraded model performance~\citep{gama2014survey}. To address this, production systems deploy drift detectors~\citep{quionero2009dataset,lu2018learning} that compare incoming data with a reference distribution and trigger alerts when differences are detected. However, most detectors are evaluated in research settings~\citep{rabanser2019failing} using synthetic shifts and single-shot tests, with emphasis on detection accuracy. In production, monitoring is continuous, models rely on many features, and alerts are generated repeatedly over time~\citep{montiel2018adaptive}. Under these conditions, even detectors with nominal false positive rates can produce frequent alerts, leading to alarm fatigue, where engineers begin to ignore or disable monitoring signals~\citep{sculley2015hidden}. Despite its practical importance, this issue is rarely studied systematically. In this work, we aim to quantify how often drift detectors ``cry wolf'' under realistic continuous monitoring and understand the trade-offs between sensitivity and stability.

We conduct controlled experiments across five widely used detectors---PSI, KS, MMD, LSDD, and adversarial validation---using the Adult Income dataset under both no-drift and gradual-drift scenarios~\citep{gama2014survey,rabanser2019failing}. We simulate 30-day monitoring cycles across multiple batch sizes and drift magnitudes, and evaluate false alarm rates, true positive rates, and time to detection~\citep{bifet2018machine,lu2018learning}. Our results confirm that PSI is highly sensitive to small batch sizes, producing frequent false alarms below approximately 200 samples~\citep{sidhu2021population,baesens2015credit}, while its false positive behavior stabilizes noticeably at larger batch sizes. In contrast, statistical detectors exhibit more consistent performance in low-data regimes but show moderate and persistent false alarm fluctuations across batch sizes under standard thresholds~\citep{benjamini1995controlling,montiel2018adaptive}. We further observe that applying a Bonferroni correction substantially reduces false alarms, consistent with prior work, but often at the expense of detection sensitivity, highlighting a clear stability--sensitivity trade-off across methods~\citep{dunn1961multiple,rabanser2019failing}. These findings provide practical deployment guidelines and highlight configuration choices that materially affect monitoring reliability in production ML systems~\citep{sculley2015hidden}.

\section{Related Work}

\subsection{Drift Detection Methods}

A wide range of drift detection methods have been proposed for monitoring distributional changes~\citep{gama2014survey,lu2018learning}. Classical statistical tests such as the Kolmogorov--Smirnov (KS) test~\citep{kolmogorov1933empirical} and Population Stability Index (PSI)~\citep{lewis1994introduction} compare univariate feature distributions and are widely used in credit scoring and tabular ML pipelines~\citep{baesens2015credit,sidhu2021population}. Kernel-based methods extend this to multivariate settings by measuring distances in reproducing kernel Hilbert spaces, with Maximum Mean Discrepancy (MMD)~\citep{gretton2012kernel} and Least-Squares Density Difference (LSDD)~\citep{sugiyama2012density} as representative examples. Adversarial detectors~\citep{lopezpaz2016revisiting}, where a classifier distinguishes reference from current data, have also gained popularity in production systems~\citep{lipton2018detecting}. These statistical, kernel, and adversarial families form the core toolkit for modern ML monitoring~\citep{rabanser2019failing}.

\subsection{Continuous Monitoring and Multiple Testing}

Most drift detection research evaluates detectors on benchmark datasets or synthetic shifts, focusing on detection accuracy and time-to-detection at single change points~\citep{bifet2018machine,montiel2018adaptive}. However, production monitoring involves continuous testing across many features and time windows~\citep{sculley2015hidden}. Under such conditions, even detectors with nominal false positive rates can generate frequent alerts due to random variation---a problem that has received limited empirical attention in the drift detection literature~\citep{rabanser2019failing}. The issue of repeated hypothesis testing is well studied in statistics under multiple testing frameworks~\citep{benjamini1995controlling}. Methods like Bonferroni correction adjust significance thresholds to control family-wise error rates, though at the cost of reduced statistical power~\citep{dunn1961multiple}. While standard in genomics and A/B testing, such corrections are rarely incorporated into drift detection pipelines, where detectors typically use default thresholds designed for single-shot experiments rather than continuous monitoring~\citep{fanaee2021survey}.

\section{Experiments}

\subsection{Experimental Setup}

We evaluate five drift detection methods on the Adult Income dataset~\citep{Dua2019}, a standard tabular benchmark with mixed numerical and categorical features~\citep{kohavi1996scaling}. A continuous monitoring scenario spanning 30 days is constructed using a fixed reference dataset and daily batches representing production data~\citep{sculley2015hidden}. In the no-drift setting, daily batches are sampled from the reference distribution. In the drift setting, controlled shifts are injected into a single feature to simulate gradual distributional changes~\citep{gama2014survey}.

\subsection{Detectors and Methods}

Five detection methods spanning different methodological families are considered: Population Stability Index (PSI), a histogram-based univariate metric widely used in industry; Kolmogorov--Smirnov (KS) test, a classical non-parametric statistical test; two kernel-based multivariate detectors, Maximum Mean Discrepancy (MMD) and Least-Squares Density Difference (LSDD); and adversarial validation, where a classifier distinguishes reference from current data. Detector behavior is examined under two scenarios: (1) batch-size sweep from 50 to 500 samples (increments of 50), and (2) drift-span experiments with shift magnitudes of 5, 10, 15, and 20 years injected into the age feature. For statistical detectors, both standard and Bonferroni-corrected thresholds are evaluated to analyze multiple-testing correction effects. Each configuration is repeated across five random seeds.

\subsection{Evaluation Metrics}

Performance is evaluated using three operational metrics: (1) false positive days---the number of days (out of 30) with alarms in the no-drift scenario~\citep{basseville1993detection}, (2) true positive rate (TPR)~\citep{gama2004learning}---the fraction of drift-period days with correct detection, and (3) time to detection (TTD)---days between drift onset and first detection~\citep{bifet2018machine}. Results are reported as mean and standard deviation across seeds.

Full experimental details are provided in Appendix~\ref{appendix:A}.

\subsection{Results}

Across all detectors, continuous monitoring reveals that false alarms are not a theoretical concern but a practical operational issue. In the no-drift setting, several detectors triggered alerts on multiple days despite the absence of any distributional change, confirming the presence of a ``cry wolf'' effect under realistic monitoring conditions.

PSI exhibited the most pronounced batch-size dependence, with alarm rates approaching daily triggers when fewer than approximately 200 samples were available, followed by a sharp reduction and stabilization as batch sizes increased. By comparison, KS and LSDD behaved more robustly in low-data settings but continued to show non-trivial false alarm variability across the full range of batch sizes. Adversarial validation demonstrated comparatively stable behavior, though with limited responsiveness to smaller distributional shifts. The introduction of a Bonferroni correction consistently suppressed false alarms for statistical detectors, but this came with a reduction in sensitivity, emphasizing the inherent stability--sensitivity trade-off across detectors and configurations.

More exploration on each detector is given in Table~\ref{summary-detector-table}. Detailed quantitative results for all settings are provided in Appendix~\ref{appendix:B}.


\begin{figure}[tbp]
  \centering
  \includegraphics[width=\linewidth]{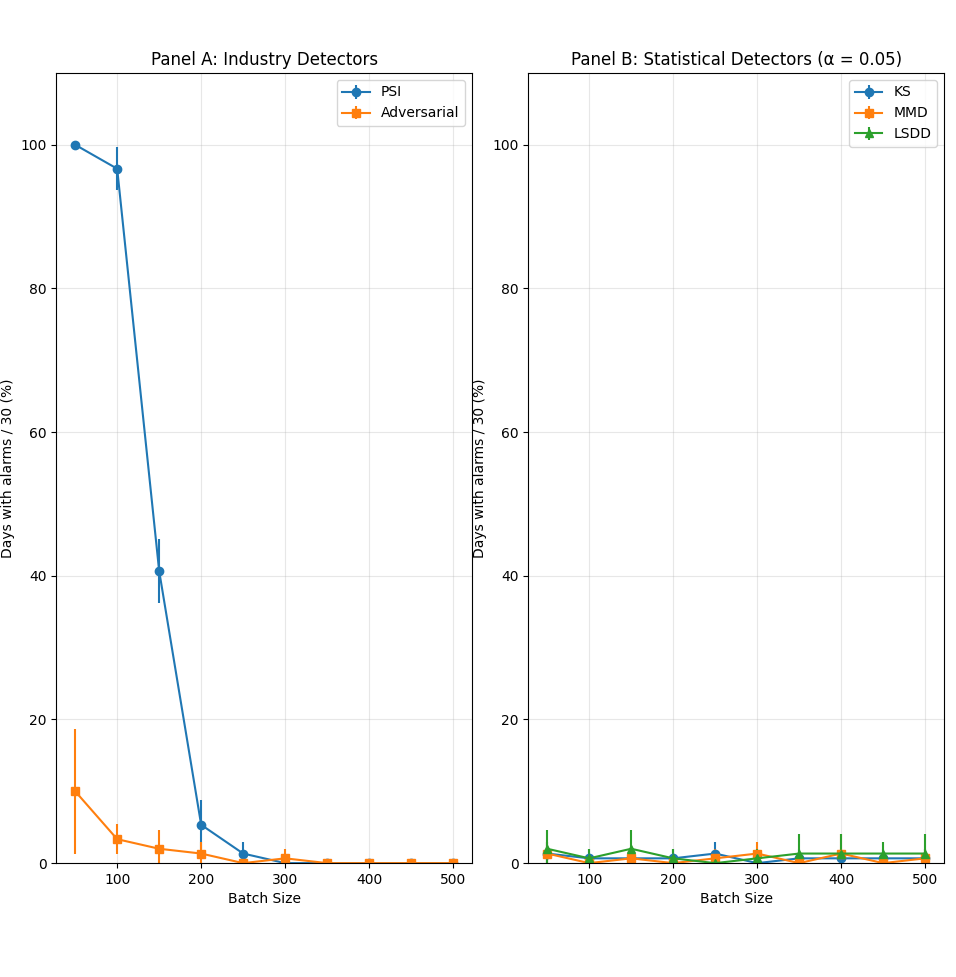}
  \caption{\textbf{False alarm rates as a function of batch size.}
    \textit{Panel A (left) — Industry-style detectors} (PSI and adversarial validation).
    PSI exhibits extreme false alarm rates at small batch sizes, approaching alarms on nearly
    all 30 days below 200 samples, followed by a sharp stability transition.
    \textit{Panel B (right) — Statistical detectors} (KS, MMD, LSDD) at $\alpha = 0.05$.
    These methods maintain relatively low false alarm rates across all batch sizes,
    indicating greater stability under continuous monitoring.}
  \label{fig:fp_batchsize}
\end{figure}


\begin{figure}[tbp]
  \centering
  \begin{minipage}[t]{0.47\linewidth}
    \centering
    \includegraphics[width=\linewidth]{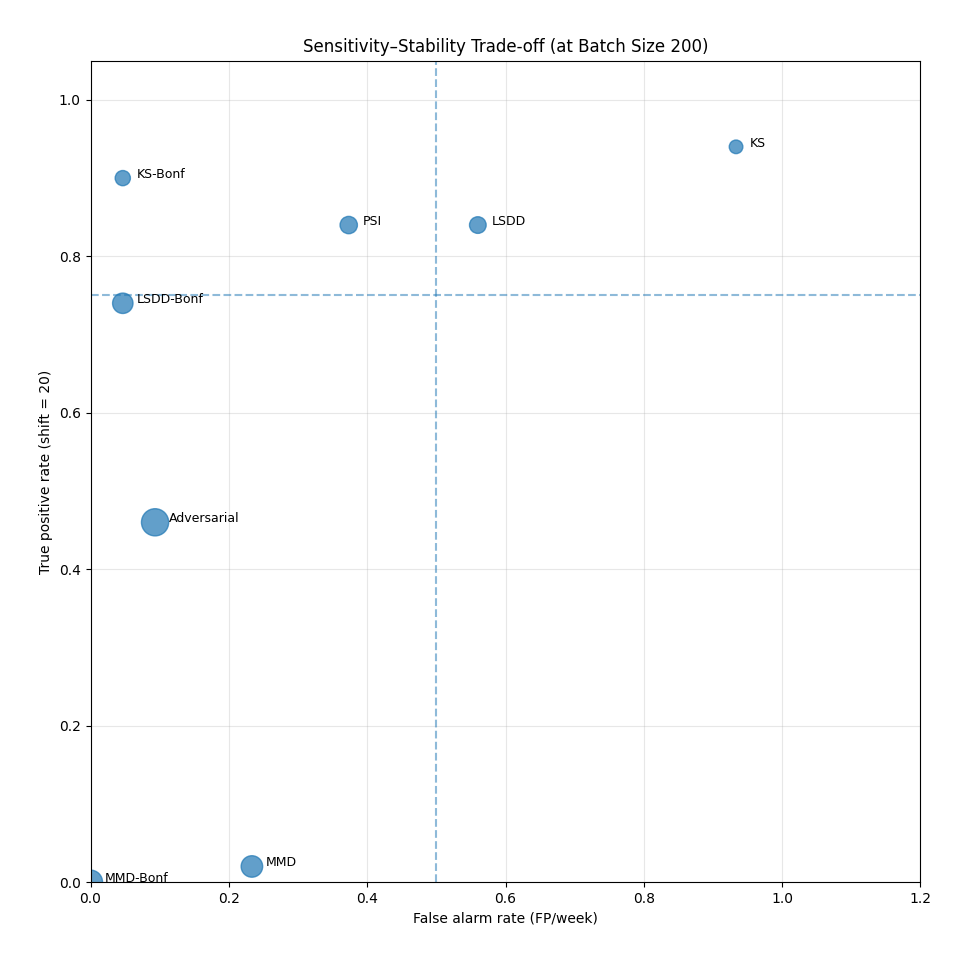}
    \caption{\textbf{Sensitivity--stability trade-off at batch size 200.}
      Each detector is plotted by its false alarm rate (x-axis) against its true positive
      rate for a strong drift of 20 years (y-axis). Detectors in the upper-left quadrant
      (low false alarms, high TPR) are most desirable. KS achieves the best overall
      balance, while MMD shows near-zero sensitivity despite low false alarms.}
    \label{fig:tradeoff}
  \end{minipage}
  \hfill
  \begin{minipage}[t]{0.47\linewidth}
    \centering
    \includegraphics[width=\linewidth]{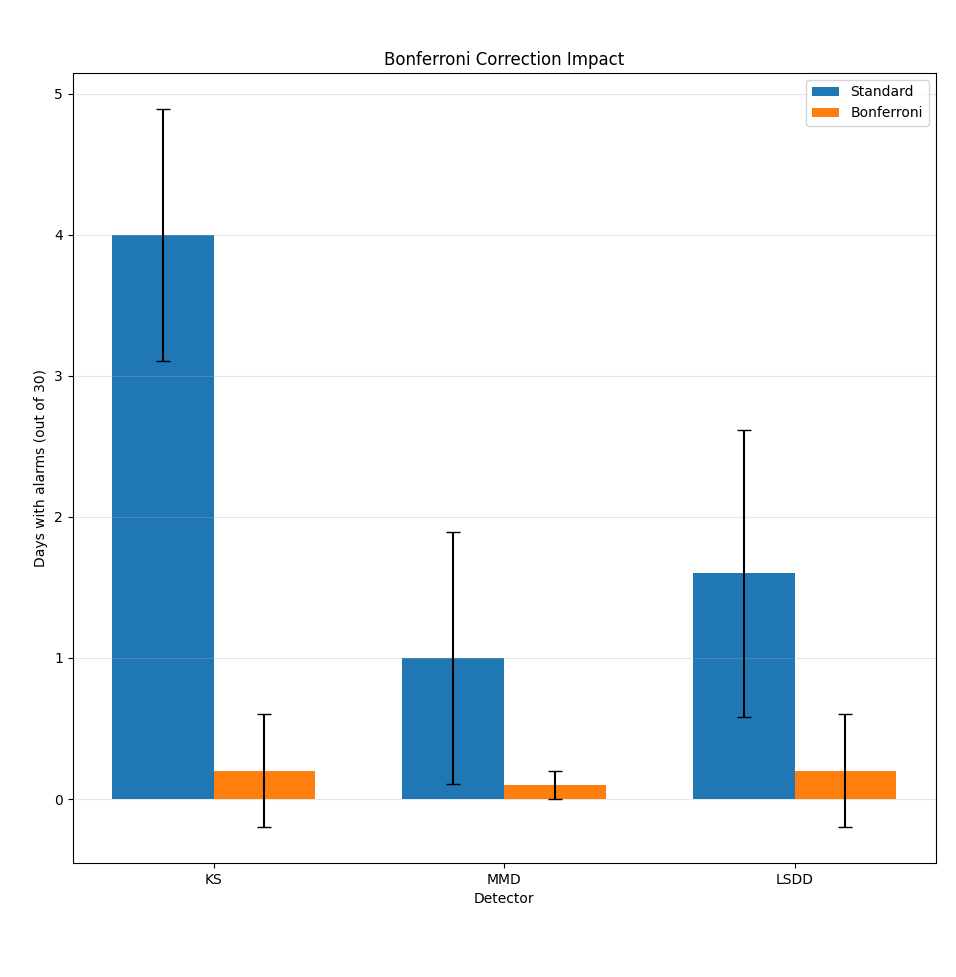}
    \caption{\textbf{Effect of Bonferroni correction on false alarms.}
      Standard vs.\ Bonferroni-corrected false alarm counts (out of 30 days) for KS,
      MMD, and LSDD at batch size 200. The correction substantially reduces false
      alarms in all three detectors, but at the cost of reduced detection sensitivity
      (see Table~\ref{summary-detector-table}).}
    \label{fig:bonferroni}
  \end{minipage}
\end{figure}


\begin{table}[tbp]
\caption{Summary of detector behavior and practical implications under continuous monitoring. Interpretation is based on experiments as well as literature review.}
\label{summary-detector-table}
\centering
\renewcommand{\arraystretch}{1.3}
\begin{tabular}{p{2.0cm} p{4.0cm} p{4.0cm} p{4.0cm}}
\toprule
\textbf{Detector} & \textbf{Key Observation} & \textbf{Interpretation} & \textbf{Practical Implication} \\
\midrule
PSI &
Extremely high false alarm rates at small batch sizes ($\approx$30/30 days at batch 50--100), sharp drop after $\sim$200 samples, near-zero thereafter &
Highly sensitive to sampling noise from small batches due to histogram-based formulation~\citep{khademi2023model,yurdakul2020statistical} &
Suitable only when batch sizes exceed $\sim$200 samples; below this threshold it produces near-constant alarms \\

KS &
Consistently low false alarms and strong detection as drift magnitude increases (TPR up to 0.90) &
Favorable balance between stability and sensitivity under standard thresholds~\citep{tonguz2025using} &
Reliable default detector for tabular monitoring with small to moderate batch sizes \\

LSDD &
Moderate false alarms and lower sensitivity than KS for small shifts &
More conservative than KS, trading sensitivity for stability under simple univariate drift~\citep{bu2017incremental} &
Suitable when fewer false alarms are preferred, at the cost of slower or weaker detection \\

MMD &
Very low false alarms but near-zero detection rates under all tested drift scenarios &
Low sensitivity to injected univariate drift under default kernel settings~\citep{zhou2025dual,schrab2023mmdagg} &
Requires careful kernel selection; default settings may miss simple localized drift \\

Adversarial &
False alarms decrease rapidly with batch size; detection only for larger shifts (TPR up to 0.46) &
Conservative: detects only strongly separable distribution changes~\citep{palli2022experimental} &
Best when minimizing false alarms is critical, though subtle drift may go undetected \\

Bonferroni correction &
Substantial reduction in false alarms (often to near-zero), accompanied by a drop in detection rates &
Stricter thresholds reduce Type I errors but lower sensitivity to small or gradual drift~\citep{liu2023concept} &
Effective when operational stability is the priority; practitioners must accept reduced sensitivity \\
\bottomrule
\end{tabular}
\end{table}

\section{Conclusion}

This work examined the behavior of commonly used drift detectors under realistic continuous monitoring conditions. While such detectors are typically evaluated on benchmark shifts with an emphasis on detection accuracy, the experiments show that their behavior can differ substantially in production-like settings. Across five detectors and a range of batch sizes, several methods produced frequent alarms even in the absence of drift, revealing a practical ``cry wolf'' problem. In particular, PSI was highly sensitive to small batch sizes, producing near-constant alarms below roughly 200 samples.

We also analyzed the trade-off between false alarms and detection sensitivity. Statistical detectors such as KS and LSDD achieved strong true positive rates but still generated non-trivial false alarm frequencies under standard thresholds. Applying a simple Bonferroni correction substantially reduced false positives, though at the cost of reduced sensitivity for some detectors, highlighting a clear sensitivity--stability trade-off across methods and configurations.

Overall, these results suggest that drift detector behavior in continuous monitoring scenarios requires careful calibration, and that standard research settings may not fully reflect production realities. Future work can extend this analysis to additional datasets, more complex drift patterns, alternative detector configurations, and longer monitoring horizons to further validate the generality of these findings.

\bibliographystyle{unsrt}
\bibliography{references}

\newpage
\appendix

\section{Appendix A: Experimental Details}
\label{appendix:A}

\subsection{Dataset Preprocessing}

\begin{table}[htbp]
\centering
\caption{Dataset preprocessing summary.}
\begin{tabular}{ll}
\toprule
\textbf{Item} & \textbf{Detail} \\
\midrule
Dataset            & Adult Income \\
Source             & UCI Machine Learning Repository \\
Missing values     & Removed \\
Categorical enc.   & Label encoding \\
Features used      & All except target \\
Final size         & $\sim$30{,}000 samples \\
Number of features & 14 \\
\bottomrule
\end{tabular}
\end{table}

\subsection{Monitoring Protocol}

\begin{table}[htbp]
\centering
\caption{Monitoring protocol summary.}
\begin{tabular}{ll}
\toprule
\textbf{Item} & \textbf{Detail} \\
\midrule
Monitoring duration   & 30 days \\
Daily execution freq. & Once per day \\
Reference dataset     & Fixed training distribution \\
Daily batch sampling  & Random sampling from dataset \\
\bottomrule
\end{tabular}
\end{table}

\noindent\textbf{No-drift experiment:}
\begin{itemize}
  \item Daily batches drawn from reference distribution
  \item Used for false positive evaluation
\end{itemize}

\noindent\textbf{Drift-span experiment:}
\begin{itemize}
  \item Drift injected into the \textit{age} feature
  \item Drift begins after day 30; gradual drift over 10 days
\end{itemize}

\subsection{Experimental Variables}

\noindent\textbf{Batch size sweep} --- Batch sizes evaluated:
\[ 50,\ 100,\ 150,\ 200,\ 250,\ 300,\ 350,\ 400,\ 450,\ 500 \]

\noindent\textbf{Drift span shifts} --- Injected shifts in age feature:
\[ 5,\ 10,\ 15,\ 20 \text{ years} \]

\noindent\textbf{Random seeds} --- All experiments repeated with seeds $\{0, 1, 2, 3, 4\}$.
Results reported as mean $\pm$ standard deviation.

\subsection{Detector Configurations}

\noindent\textbf{PSI:}
\begin{itemize}
  \item Binning: 10 equal-frequency bins
  \item Threshold: 0.2 (standard industry setting)
\end{itemize}

\noindent\textbf{KS Test:}
\begin{itemize}
  \item Two-sample KS test; standard threshold $\alpha = 0.05$
  \item Bonferroni-corrected: $\alpha = 0.05\,/\,\text{num\_features} \approx 0.007$ (14 features)
\end{itemize}

\noindent\textbf{MMD:}
\begin{itemize}
  \item Gaussian kernel (default); reference subset: 2000 samples; permutations: 100
  \item Standard $\alpha = 0.05$; Bonferroni $\alpha \approx 0.007$
\end{itemize}

\noindent\textbf{LSDD:}
\begin{itemize}
  \item Gaussian kernel (default); reference subset: 2000 samples; permutations: 100
  \item Standard $\alpha = 0.05$; Bonferroni $\alpha \approx 0.007$
\end{itemize}

\noindent\textbf{Adversarial Detector:}
\begin{itemize}
  \item Model: logistic regression; train/test split: reference vs.\ current batch
  \item Metric: ROC-AUC; alarm threshold: AUC $\geq 0.6$
\end{itemize}

\subsection{Evaluation Metrics}

\begin{table}[htbp]
\centering
\caption{Evaluation metrics summary.}
\begin{tabular}{lll}
\toprule
\textbf{Metric} & \textbf{Definition} & \textbf{Notes} \\
\midrule
False positive days  & Days with alarm in no-drift setting        & Range: 0--30 \\
True positive rate   & Fraction of drift-period days with alarm   & Drift window only \\
Time to detection    & Days from drift start to first detection   & Mean $\pm$ std across seeds \\
\bottomrule
\end{tabular}
\end{table}

\subsection{Implementation Details}

\begin{table}[htbp]
\centering
\caption{Implementation details.}
\begin{tabular}{ll}
\toprule
\textbf{Item} & \textbf{Detail} \\
\midrule
Language  & Python \\
Libraries & NumPy, Pandas, scikit-learn, alibi-detect, evidently \\
Hardware  & CPU-based experiments \\
RNG       & NumPy default RNG \\
\bottomrule
\end{tabular}
\end{table}

\section{Appendix B: Complete Experimental Tables}
\label{appendix:B}

\subsection*{False Positives vs.\ Batch Size (No-Drift Scenario)}

\noindent False positives = number of alarm days out of 30, reported as mean $\pm$ std across 5 seeds.

\vspace{0.6em}
\begin{table}[htbp]
\centering
\caption{False positives vs.\ batch size (50--200).}
\begin{tabular}{llcccc}
\toprule
\textbf{Detector} & \textbf{Variant} & \textbf{50} & \textbf{100} & \textbf{150} & \textbf{200} \\
\midrule
Adversarial & Standard   & 3.00$\pm$2.61  & 1.00$\pm$0.63  & 0.60$\pm$0.80  & 0.40$\pm$0.49 \\
KS          & Standard   & 0.40$\pm$0.49  & 0.20$\pm$0.40  & 0.20$\pm$0.40  & 0.20$\pm$0.40 \\
KS          & Bonferroni & 0.40$\pm$0.49  & 0.20$\pm$0.40  & 0.20$\pm$0.40  & 0.20$\pm$0.40 \\
LSDD        & Standard   & 0.60$\pm$0.80  & 0.20$\pm$0.40  & 0.60$\pm$0.80  & 0.20$\pm$0.40 \\
LSDD        & Bonferroni & 0.60$\pm$0.80  & 0.20$\pm$0.40  & 0.60$\pm$0.80  & 0.20$\pm$0.40 \\
MMD         & Standard   & 0.40$\pm$0.49  & 0.00$\pm$0.00  & 0.20$\pm$0.40  & 0.00$\pm$0.00 \\
MMD         & Bonferroni & 0.40$\pm$0.49  & 0.00$\pm$0.00  & 0.20$\pm$0.40  & 0.00$\pm$0.00 \\
PSI         & Standard   & 30.00$\pm$0.00 & 29.00$\pm$0.89 & 12.20$\pm$1.33 & 1.60$\pm$1.02 \\
\bottomrule
\end{tabular}
\end{table}

\begin{table}[htbp]
\centering
\caption{False positives vs.\ batch size (250--500).}
\begin{tabular}{llcccccc}
\toprule
\textbf{Detector} & \textbf{Variant} & \textbf{250} & \textbf{300} & \textbf{350} & \textbf{400} & \textbf{450} & \textbf{500} \\
\midrule
Adversarial & Standard   & 0.00$\pm$0.00 & 0.20$\pm$0.40 & 0.00$\pm$0.00 & 0.00$\pm$0.00 & 0.00$\pm$0.00 & 0.00$\pm$0.00 \\
KS          & Standard   & 0.40$\pm$0.49 & 0.00$\pm$0.00 & 0.20$\pm$0.40 & 0.20$\pm$0.40 & 0.20$\pm$0.40 & 0.20$\pm$0.40 \\
KS          & Bonferroni & 0.40$\pm$0.49 & 0.00$\pm$0.00 & 0.20$\pm$0.40 & 0.20$\pm$0.40 & 0.20$\pm$0.40 & 0.20$\pm$0.40 \\
LSDD        & Standard   & 0.00$\pm$0.00 & 0.20$\pm$0.40 & 0.40$\pm$0.80 & 0.40$\pm$0.80 & 0.40$\pm$0.49 & 0.40$\pm$0.80 \\
LSDD        & Bonferroni & 0.00$\pm$0.00 & 0.20$\pm$0.40 & 0.40$\pm$0.80 & 0.40$\pm$0.80 & 0.40$\pm$0.49 & 0.40$\pm$0.80 \\
MMD         & Standard   & 0.20$\pm$0.40 & 0.40$\pm$0.49 & 0.00$\pm$0.00 & 0.40$\pm$0.80 & 0.00$\pm$0.00 & 0.20$\pm$0.40 \\
MMD         & Bonferroni & 0.20$\pm$0.40 & 0.40$\pm$0.49 & 0.00$\pm$0.00 & 0.40$\pm$0.80 & 0.00$\pm$0.00 & 0.20$\pm$0.40 \\
PSI         & Standard   & 0.40$\pm$0.49 & 0.00$\pm$0.00 & 0.00$\pm$0.00 & 0.00$\pm$0.00 & 0.00$\pm$0.00 & 0.00$\pm$0.00 \\
\bottomrule
\end{tabular}
\end{table}

\vspace{1em}
\noindent\rule{\linewidth}{0.4pt}
\vspace{0.5em}

\subsection*{True Positive Performance vs.\ Drift Magnitude}

\noindent Metrics reported as mean $\pm$ std across 5 seeds.

\vspace{0.6em}
\begin{table}[htbp]
\centering
\caption{True positive performance vs.\ drift magnitude.}
\begin{tabular}{llcccc}
\toprule
\textbf{Detector} & \textbf{Variant} & \textbf{Shift} & \textbf{TPR} & \textbf{FP} & \textbf{TTD (days)} \\
\midrule
Adversarial & Standard & 5  & 0.00$\pm$0.00 & 0.00$\pm$0.00 & N/A \\
Adversarial & Standard & 10 & 0.00$\pm$0.00 & 0.00$\pm$0.00 & N/A \\
Adversarial & Standard & 15 & 0.28$\pm$0.04 & 0.00$\pm$0.00 & 8.20$\pm$0.40 \\
Adversarial & Standard & 20 & 0.46$\pm$0.05 & 0.00$\pm$0.00 & 6.40$\pm$0.49 \\
\midrule
KS & Standard & 5  & 0.30$\pm$0.00 & 0.20$\pm$0.40 & 7.40$\pm$1.20 \\
KS & Standard & 10 & 0.68$\pm$0.04 & 0.20$\pm$0.40 & 4.20$\pm$0.40 \\
KS & Standard & 15 & 0.86$\pm$0.05 & 0.20$\pm$0.40 & 2.20$\pm$0.40 \\
KS & Standard & 20 & 0.90$\pm$0.06 & 0.20$\pm$0.40 & 2.00$\pm$0.63 \\
\midrule
LSDD & Standard & 5  & 0.08$\pm$0.07 & 0.40$\pm$0.49 & 7.00$\pm$0.00 \\
LSDD & Standard & 10 & 0.48$\pm$0.07 & 0.40$\pm$0.49 & 5.80$\pm$1.17 \\
LSDD & Standard & 15 & 0.64$\pm$0.08 & 0.40$\pm$0.49 & 4.40$\pm$0.80 \\
LSDD & Standard & 20 & 0.74$\pm$0.05 & 0.20$\pm$0.40 & 3.60$\pm$0.49 \\
\midrule
MMD & Standard & 5  & 0.00$\pm$0.00 & 0.40$\pm$0.49 & N/A \\
MMD & Standard & 10 & 0.00$\pm$0.00 & 0.40$\pm$0.49 & N/A \\
MMD & Standard & 15 & 0.00$\pm$0.00 & 0.20$\pm$0.40 & N/A \\
MMD & Standard & 20 & 0.00$\pm$0.00 & 0.00$\pm$0.00 & N/A \\
\midrule
PSI & Standard & 5  & 0.40$\pm$0.06 & 1.60$\pm$1.02 & 5.20$\pm$2.14 \\
PSI & Standard & 10 & 0.66$\pm$0.05 & 1.60$\pm$1.02 & 3.60$\pm$0.80 \\
PSI & Standard & 15 & 0.74$\pm$0.05 & 1.60$\pm$1.02 & 3.20$\pm$0.98 \\
PSI & Standard & 20 & 0.84$\pm$0.05 & 1.60$\pm$1.02 & 2.60$\pm$0.49 \\
\bottomrule
\end{tabular}
\end{table}

\end{document}